\documentclass{llncs}

\usepackage{graphicx,color,soul,array,amsmath,amssymb}
\usepackage{dvmacros}

\begin{document}


\newcommand{\weightdecay}{10^{-4}}

\newcommand{\CNN}{CNN}
\newcommand{\FT}{FT}
\newcommand{\FTCMR}{FT-CMR}
\newcommand{\IoU}{IoU}
\newcommand{\CRF}{CRF}
\newcommand{\SGD}{SGD}
\newcommand{\networkmomentum}{0.9}
\newcommand{\UNet}{U-Net}
\newcommand{\DC}{DC}

\newcommand{\N}{128}

\newcommand{\BLVEndC}{-26.3}
\newcommand{\BLVEndO}{-29.0}
\newcommand{\BLVEndp}{0.098}
\newcommand{\BLVEpiC}{-11.8}
\newcommand{\BLVEpiO}{-11.7}
\newcommand{\BLVEpip}{0.952}
\newcommand{\BRVEndC}{-11.9}
\newcommand{\BRVEndO}{-14.3}
\newcommand{\BRVEndp}{0.529}
\newcommand{\ELVEndC}{-25.3}
\newcommand{\ELVEndO}{-29.1}
\newcommand{\ELVEndp}{0.006}
\newcommand{\ELVEpiC}{-8.8}
\newcommand{\ELVEpiO}{-8.5}
\newcommand{\ELVEpip}{0.661}
\newcommand{\ERVEndC}{-8.0}
\newcommand{\ERVEndO}{-12.7}
\newcommand{\ERVEndp}{0.152}
\newcommand{\ALVEndC}{-25.5}
\newcommand{\ALVEndO}{-28.6}
\newcommand{\ALVEndp}{0.110}
\newcommand{\ALVEpiC}{-10.7}
\newcommand{\ALVEpiO}{-9.9}
\newcommand{\ALVEpip}{0.470}
\newcommand{\ARVEndC}{-13.8}
\newcommand{\ARVEndO}{-12.4}
\newcommand{\ARVEndp}{0.566}

\newcommand{\NumC}{21}
\newcommand{\NumO}{42}
\newcommand{\NumFolds}{3}

\newcommand{\ChOneAtFAcc}{0.977}
\newcommand{\ChOneAtFIoUAll}{0.874}
\newcommand{\ChOneAtFIoUFor}{0.838}
\newcommand{\ChOneAtTAcc}{0.977}
\newcommand{\ChOneAtTIoUAll}{0.876}
\newcommand{\ChOneAtTIoUFor}{0.840}
\newcommand{\ChTwoAtFAcc}{0.976}
\newcommand{\ChTwoAtFIoUAll}{0.874}
\newcommand{\ChTwoAtFIoUFor}{0.838}
\newcommand{\ChTwoAtTAcc}{0.976}
\newcommand{\ChTwoAtTIoUAll}{0.873}
\newcommand{\ChTwoAtTIoUFor}{0.837}

\newcommand{\ChOneAtTAccMed   }{0.981}
\newcommand{\ChOneAtTIoUAllMed}{0.886}
\newcommand{\ChOneAtTIoUForMed}{0.853}
\newcommand{\ChTwoAtTAccMed   }{0.981}
\newcommand{\ChTwoAtTIoUAllMed}{0.888}
\newcommand{\ChTwoAtTIoUForMed}{0.855}
\newcommand{\ChOneAtFAccMed   }{0.981}
\newcommand{\ChOneAtFIoUAllMed}{0.885}
\newcommand{\ChOneAtFIoUForMed}{0.851}
\newcommand{\ChTwoAtFAccMed   }{0.981}
\newcommand{\ChTwoAtFIoUAllMed}{0.887}
\newcommand{\ChTwoAtFIoUForMed}{0.855}


\title{Feature Tracking Cardiac Magnetic Resonance via Deep Learning and Spline Optimization}
\titlerunning{Feature Tracking CMR}

\author{
Davis M. Vigneault\inst{1,2,3} \and
Weidi Xie\inst{1} \and
David A. Bluemke\inst{2} \and
J. Alison Noble\inst{1}
}

\authorrunning{Davis M. Vigneault et al.} 

\institute{
Institute of Biomedical Engineering, Department of Engineering, University of Oxford, Old Road Campus Research Building, Roosevelt Dr, Oxford, United Kingdom, OX3 7DQ,\\
\email{davis.vigneault@gmail.com},\\
\and
Department of Radiology and Imaging Sciences, Clinical Center, National Institutes of Health, 10 Center Drive, Bethesda, MD, United States, 20814.
\and
Sackler School of Graduate Biomedical Sciences, Tufts University School of Medicine, 136 Harrison Ave, Boston, MA, United States, 02111.
}

\maketitle


\begin{abstract} 
Feature tracking Cardiac Magnetic Resonance (\CMR{}) has recently emerged as an area of interest for quantification of regional cardiac function from balanced, steady state free precession (\SSFP{}) cine sequences.  However, currently available techniques lack full automation, limiting reproducibility.  We propose a fully automated technique whereby a \CMR{} image sequence is first segmented with a deep, fully convolutional neural network (\CNN{}) architecture, and quadratic basis splines are fitted simultaneously across all cardiac frames using least squares optimization.  Experiments are performed using data from \NumO{} patients with hypertrophic cardiomyopathy (\HCM{}) and \NumC{} healthy control subjects.  In terms of segmentation, we compared state-of-the-art \CNN{} frameworks, \UNet{} and dilated convolution architectures, with and without temporal context, using cross validation with three folds.  Performance relative to expert manual segmentation was similar across all networks: pixel accuracy was $\sim 97\%$, intersection-over-union (\IoU{}) across all classes was $\sim 87\%$, and \IoU{} across foreground classes only was $\sim 85\%$.  Endocardial left ventricular circumferential strain calculated from the proposed pipeline was significantly different in control and disease subjects ($\ELVEndC{}\%$ vs $\ELVEndO{}\%$, $p=\ELVEndp{}$), in agreement with the current clinical literature.

\end{abstract}


\section{Introduction}

\begin{figure}[htb]
\centering
Match Successive Pairs of Frames
\includegraphics[clip, trim=25 5 10 5, width=\textwidth]{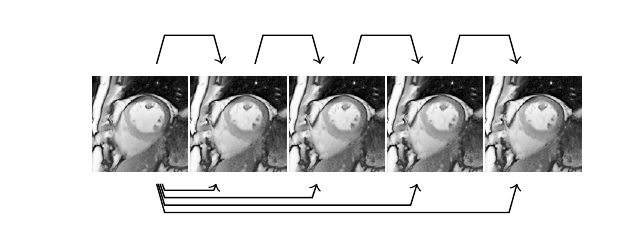}
Match Each Frame to End Diastole
\caption{Schematic of traditional feature tracking methods.  Each arrow represents a single pairwise registration between a fixed and moving image, and produces a displacement field.  Through combinations of resampling, averaging, and regularization, these displacement fields are combined to form a final sequence of fields representing cardiac motion.}
\label{fig:pairwisetracking}
\end{figure}

Quantification of regional cardiac function is of utmost importance in the characterization of subtle abnormalities which may precede changes in global metrics \cite{Tee2013}.  Harmonic phase (\HARP{}) analysis \cite{Osman2000} of tagged cardiac magnetic resonance (\CMR{}) images is the gold-standard for regional function, but has not been adopted clinically due to lengthy acquisition and analysis.  Moreover, \HARP{} analysis is difficult to apply to chambers other than the left ventricle (\LV{}) due to the thinness of the myocardial wall in the atria and right ventricle (\RV{}).  Recently, feature tracking (\FT{}) has emerged as a promising alternative to tagging \cite{Pedrizzetti2016}.  Because \FTCMR{} can be applied to balanced steady state free precession (\SSFP{}) images, acquisition of specialized image sequences is avoided.  Moreover, because \FTCMR{} primarily tracks myocardial borders and trabeculation, the thinness of the atrial and right ventricular myocardium does not hinder tracking.

Despite recent interest in \FTCMR{}, two important challenges remain.  First, all current commercially available implementations (MTT, TomTec, CMR42) require manual contouring of one or more cine frames, preventing full automation and reducing reproducibility.  Second, \FT{} generally has been implemented by repeatedly applying methods designed to determine a displacement field between a \emph{single pair} of images (e.g., optical flow, block matching, deformable registration), rather than an image \emph{sequence}; either matching successive pairs, or matching each frame to a single reference, typically end diastole (\ED{}, Fig~\ref{fig:pairwisetracking}).  Each of these approaches has well-known potential drawbacks \cite{Wong2016}, which may be overcome by empirically optimizing over all frames simultaneously \cite{Stebbing2014}.

Here, we propose a method for \FTCMR{} analysis which overcomes the first of these challenges by using a deep learning approach in place of human contouring.  Deep convolutional neural networks have been used to great effect in image classification \cite{Krizhevsky,Simonyan2015}, and semantic segmentation \cite{Long2015,Yu2016}.  Recently, \CNN{}s have also shown state-of-the-art performance in biomedical image analysis \cite{Ronneberger2015,Xie2015}.  \CNN{} segmentation of short axis \CMR{} has been applied to the \LV{} blood-pool \cite{Poudel2016a}, the \RV{} blood-pool \cite{Luo2016}, and both simultaneously \cite{Tran2016}.  Here, we perform segmentation of the \LV{} myocardium, \LV{} blood-pool, and \RV{} blood-pool.  Moreover, we apply the segmentation to patients with \HCM{}, which increases the complexity of the problem due to the highly variable appearance of the \LV{} in these patients.

The second of these challenges we address by fitting quadratic basis splines to the segmentation data jointly, rather than frame by frame, adapting the technique presented in \cite{Stebbing2015,Stebbing2014} in the context of cardiac ultrasound.  In our pipeline, \emph{extraction} is performed using a \CNN{}, \emph{tracking} is performed with simultaneous spline optimization, and cardiac strain is estimated from the registered splines.

\section{Methods}

Broadly, the automated analysis pipeline involves three steps: feature extraction (segmentation), feature tracking (spline fitting), and calculation of functional parameters (strain estimation).  These steps are discussed in detail in the following sub-sections.

\subsection{Segmentation}

Following the work of \cite{Long2015,Ronneberger2015,Xie2015}, we designed our segmentation architecture as a fully convolutional network.  In order to obtain a segmentation map with the same spatial resolution as the input image, up-sampling operators are used to replace the pooling operators in traditional classification networks.  This strategy enables our network to segment arbitrarily large images.

\begin{figure}[htb]
\centering
\includegraphics[clip, trim=0 0 0 0,width=0.8\textwidth]{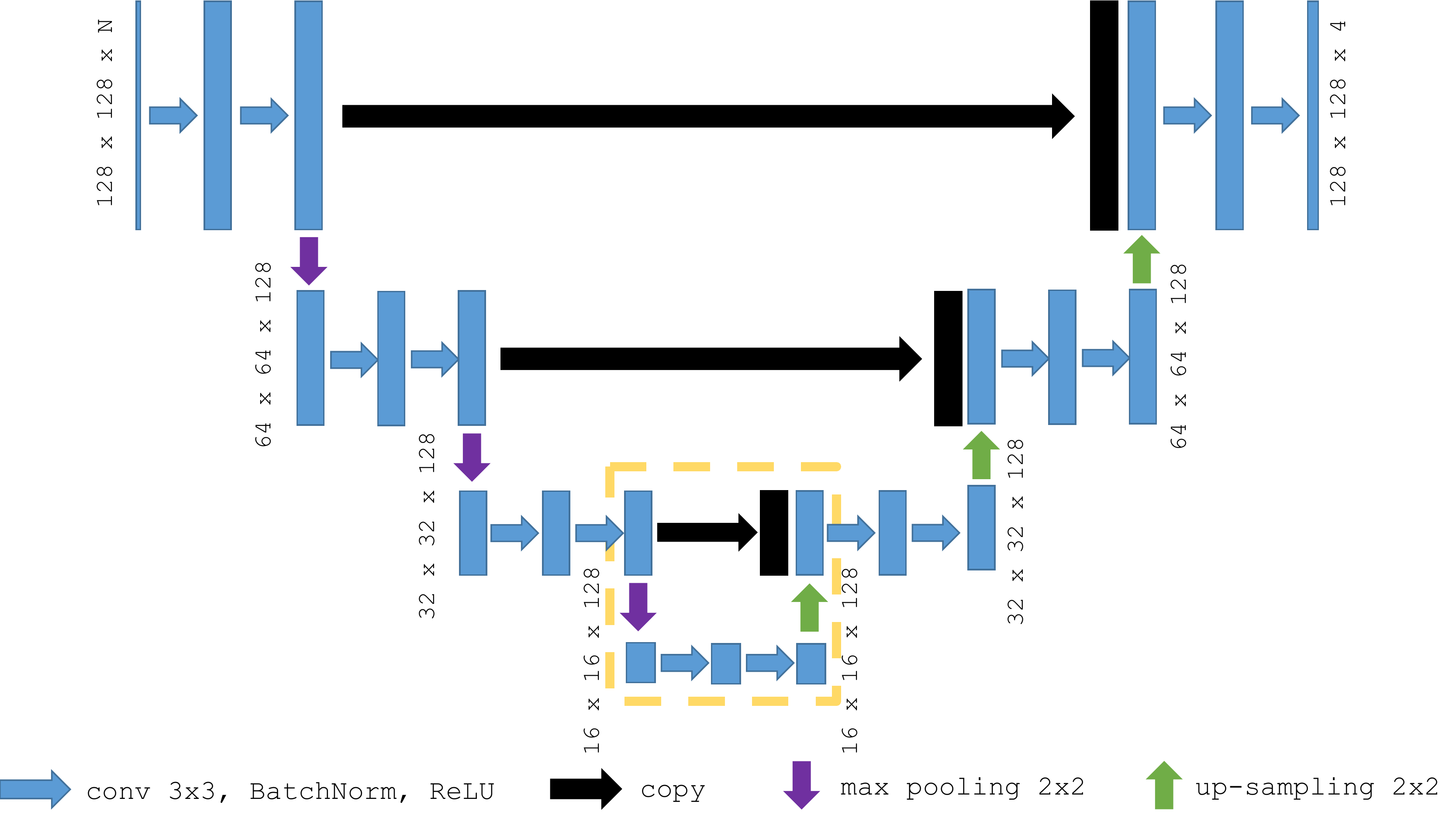}
\caption{Architecture of the basic network (Network A).  The input image is of size $\N \times \N \times N$, where $N$ is the number of channels (1 in networks A and B, 2 in networks C and D).  Each blue and black box corresponds to a multilingual feature map (black indicates the result of a copy).  The dimensions of the feature maps are indicated in the figure as first spatial dimension $\times$ second spatial dimension $\times$ channels.  The number of channels in each feature map is fixed at $128$.  The dashed yellow box is replaced by dilated convolution in networks B and D.}
\label{fig:cnnarchitecture}
\end{figure}

As shown in Fig.~\ref{fig:cnnarchitecture}, the architecture consists of a down-sampling path (left) followed by an up-sampling path (right).  During the first several layers, the structure resembles the canonical classification \CNN{} \cite{Krizhevsky,Simonyan2015}, as a $3 \times 3$ convolution, rectified linear unit (ReLU), and $2 \times 2$ max pooling are repeatedly applied to the input image and feature maps.  In the second half of the architecture, we ``undo'' the reduction in spatial resolution by performing $2 \times 2$ up-sampling, ReLU activation, and $3 \times 3$ convolution, eventually mapping the intermediate feature representation back to the original resolution.  To provide accurate boundary localization, low-level feature representations from the down-sampling path are concatenated with the feature maps from the up-sampling path.  For all layers, we apply 128 trainable kernels.  We performed batch normalization \cite{Ioffe2015}, which has been shown to increase generalizability, between each pair of convolution and ReLU activation layers.

\begin{table}
\centering
\caption{\CNN{} architecture variants considered. Note: \ED{} = End Diastole; \DC{} = Dilated Convolution.} \label{tab:architecturedescriptions}
\begin{tabular}{|l|l|l|l|} \hline
\hline
Name      & Variant & Input Size            & Temporal Context \\
\hline              
\hline              
Network A & \UNet{} & $ \N\times\N\times1 $ & None             \\
Network B & \DC{}   & $ \N\times\N\times1 $ & None             \\
                    
Network C & \UNet{} & $ \N\times\N\times2 $ & \ED{} Frame      \\
Network D & \DC{}   & $ \N\times\N\times2 $ & \ED{} Frame      \\
\hline
\end{tabular}
\end{table}

In addition to this basic architecture (Network A), we varied the amount of temporal context by either inputting the input image alone, or the input image and \ED{} image together.  We based this on the intuition that the papillary muscles, which frequently interfere with \LV{} segmentation, are least compacted at \ED{} and may guide the segmentation of the input frame.  Additionally, in Networks B and D, the final up-sampling/down-sampling pass was replaced by a dilated convolution (\DC{}).  All architectures have $\sim 3.1$ million trainable parameters.  The architectures tested are summarized in Table~\ref{tab:architecturedescriptions}.

\subsection{Quadratic \Bezier{} Curve Registration}

Following semantic segmentation, contours defining the boundaries of the \LV{} endocardium, \LV{} epicardium, and \RV{} endocardium were extracted through standard morphological operations.  In this work, the pixels belonging to these contours are known as ``boundary candidates.''  Unfortunately, these contours cannot be used directly to quantify cardiac function, because they lack anatomical correspondence between frames.  It is the aim of this section to describe an optimization procedure for jointly registering a sequence of closed, quadratic \Bezier{} curves to these boundary candidates.

A segment of a closed \Bezier{} curve $B$ of degree $d$ parameterized by $\UOne \in [0, 1]$ is a linear combination of $d+1$ control points $x_i : 0 \le i \le d$,

$$
B_d(\UOne) = \sum_{i=0}^d b_{i,d}(\UOne) x_i,
$$

\noindent where $b_{i,d}(\UOne)$ is the $i$\th{} Bernstein polynomial of degree $d$,

$$
b_{i,d}(\UOne) = \binom{n}{i} \left(1 - \UOne \right)^{n-i}r^i,
$$

\noindent and $\binom{n}{i}$, often read aloud as ``$n$ choose $i$'', is the binomial coefficient,

$$
\binom{n}{i} = \frac{n!}{i!(n-i)!}.
$$

\noindent For $d=2$, $B_2(\UOne)$ expands to

$$
B_2(\UOne) = (1-\UOne)^2 x_0 + 2\UOne(1-\UOne) x_1 + \UOne^2 x_2,
$$

\noindent which may be more conveniently expressed in terms of the monomial basis as

$$
B_2(\UOne) =
\begin{bmatrix}
x_0 & x_1 & x_2
\end{bmatrix}
\begin{bmatrix}
(1-\UOne)^2 \\
2\UOne(1-\UOne) \\
\UOne^2
\end{bmatrix} \\
=
\begin{bmatrix}
x_0 & x_1 & x_2
\end{bmatrix}
\begin{bmatrix}
1 & -2 & 1 \\
0 & 2 & -2 \\
0 & 0 & 1
\end{bmatrix}
\begin{bmatrix}
1 \\
\UOne \\
\UOne^2
\end{bmatrix}.
$$

\noindent Importantly, the first and second derivatives of $B_2(\UOne)$ with respect to $r$ are trivial to compute.

\subsection{Formulating the Optimization}

Levenberg-Marquardt least squares optimization \cite{Marquardt1963} is used to register a set of closed, quadratic \Bezier{} curves to the boundary candidates.  The parameters $\Delta X \in \Real^{2 \times \left( \NumCP \times \NumFrames \right) }$ (where $\NumCP$ is the number of control points in a single curve and $\NumFrames$ is the number of cardiac phases) of the optimization are Cartesian displacements to the control points of all template curves across all frames.  A fixed number of points $\mathbf{u}_{\FrameIndex, \PatchIndex, \UOne}$ were sampled across each curve.  At each step in the optimization, for each of these points, the nearest boundary candidate $\phi \left( \mathbf{u}_{\FrameIndex, \PatchIndex, \UOne} \right)$ was calculated, where $\phi : \Real^2 \rightarrow \Real^2$.  This was computed efficiently by representing the boundary candidate point set at each frame as a $K_d$ tree.  The Cartesian components of the distance between the points sampled from the curve and nearest boundary candidate were the residuals of $E_{cf}$, the first term of the cost function,

\begin{equation} \label{eqn:LMCF}
E_{cf} = \sum_{\FrameIndex, \PatchIndex, \UOne}
\|
\mathbf{u}_{\FrameIndex, \PatchIndex, \UOne} - \phi \left( \mathbf{u}_{\FrameIndex, \PatchIndex, \UOne} \right)
\|^2.
\end{equation}

Additionally, two regularizers were included to enforce physical constraints of anatomical deformation: control point acceleration and spline curvature.

In our cost function, the control point acceleration regularizer allows information to be shared between frames.  At a minimum, regularizing against control point velocity as in \cite{Stebbing2014} is necessary to maintain anatomical consistency (the assumption that, for fixed $\PatchIndex$ and $\UOne$, $\mathbf{u}_{\FrameIndex,\PatchIndex,\UOne}$ corresponds to the same material point $\forall \FrameIndex$).  By regularizing against acceleration rather than velocity, our method additionally encourages smooth, biologically plausible motion.  The control point acceleration regularizer $E_{ac}$ was defined as the Cartesian components of the second differences between corresponding vertices $\mathbf{x}_{\FrameIndex, \CPIndex}$ in three adjacent frames, where $\mathbf{x}_{\FrameIndex, \CPIndex}$ is the $(\FrameIndex \times \NumCP) + \CPIndex$\th{} column of $X$.

\begin{equation} \label{eqn:LMAC}
E_{ac} = \sum_{\FrameIndex, \CPIndex}
\left \lVert
\begin{bmatrix}
1 & -2 & 1
\end{bmatrix}
\begin{bmatrix}
\mathbf{x}^\top_{(\FrameIndex+2) \bmod \NumFrames,\CPIndex} \\
\mathbf{x}^\top_{(\FrameIndex+1) \bmod \NumFrames,\CPIndex} \\
\mathbf{x}^\top_{ \FrameIndex,\CPIndex}
\end{bmatrix}
\right \lVert^2
\end{equation}

Curvature for segment $\PatchIndex$ in frame $\FrameIndex$ is the second derivative of $B_2(\UOne)$ with respect to $\UOne$.

\begin{equation} \label{eqn:LMCV}
E_{cv} = \sum_{\FrameIndex,\CPIndex} \left \lVert \frac{d^2B_2(\UOne)}{dr^2} \right \lVert^2
\end{equation}

The overall optimization problem may then be written in terms of Eqns.~\ref{eqn:LMCF}, \ref{eqn:LMAC}, and \ref{eqn:LMCV} and corresponding scaling factors.  Scaling factors $\rho_{cf} = 10.0$, $\rho_{ac} = 1.0$, and $\rho_{cv} = 0.1$ were set empirically to prevent any single term from dominating the optimization.

$$
E = \min \left( \rho_{cf} E_{cf} + \rho_{ac} E_{ac} + \rho_{cv} E_{cv} \right)
$$

Two points relating to computational efficiency are worth noting.  First, the Jacobians of Eqns.~\ref{eqn:LMCF}, \ref{eqn:LMAC}, and \ref{eqn:LMCV} can all be calculated analytically.  By providing explicit Jacobians, we avoid the need for numeric derivatives, which would slow computation precipitously.  Moreover, for a given set of correspondences between surface positions and boundary candidates, the Jacobians of all residuals are linear with respect to the Cartesian displacements of the control points and therefore trivial to calculate.  Second, each individual residual depends upon a very small number of parameters.  Specifically, each individual residual depends on exactly three control points (six parameters).  This sparsity is exploited during the optimization to limit the number of components of the Jacobian which must be evaluated, further reducing computational cost.

Following the initial fit, the spline is subdivided and used to initialize a second optimization, and this process is repeated one further time.  This multiresolution approach has benefit over registering the highly subdivided spline directly, which can be sensitive to initialization.

\section{Experiments}

\subsection{Segmentation}

\newcommand{\NImFoldZero}{2706}
\newcommand{\NImFoldOne}{3000}
\newcommand{\NImFoldTwo}{2775}
\newcommand{\NumCinesTotal}{189}
\newcommand{\NumSubjectsTotal}{63}
\newcommand{\NumFramesMin}{25}
\newcommand{\NumFramesMax}{50}

The \LV{} myocardium, \LV{} blood-pool, and \RV{} blood-pool were manually segmented in \NumCinesTotal{} short axis 2D+time volumes (basal, equatorial, and apical cine series from each of \NumSubjectsTotal{} subjects).  The papillary muscles of the \LV{} were excluded from the myocardium.  The subjects were partitioned into three folds of approximately equal size such that the images from any one subject were present in one fold only.  The volumes were cropped to $\N \times \N$ pixels in the spatial dimensions, and varied from $\NumFramesMin{}$ to $\NumFramesMax{}$ pixels in the time dimension, totaling $\NImFoldZero{}$, $\NImFoldOne{}$, and $\NImFoldTwo{}$ images in the three folds, respectively.  For each of the four architectures (\UNet{} and \DC{} with and without temporal information), three models were trained on two folds and tested on the remaining fold.  The images were histogram equalized and normalized to zero mean and unit standard deviation before being input into the \CNN{}.  The network weights were initialized with orthogonal weights \cite{Saxe2013}, and were trained with standard stochastic gradient descent (\SGD{}) with momentum (\networkmomentum{}) by optimizing categorical cross entropy.  Learning rate was initialized to $0.01$ and decayed by $0.1$ every $32$ epochs.  To avoid over-fitting, we used considerable data augmentation (horizontal and vertical flipping, random translations and rotation) and a weight decay of $\weightdecay{}$.  Accuracy was measured as pixel accuracy between the prediction and manual segmentations.  The model was implemented in the Python programming language using the Keras interface to Tensorflow \cite{Barham2016}, and trained on one NVIDIA Titan X graphics processing units (GPU) with 12 GB of memory.  For all network architectures, it took roughly 200 seconds to iterate over the entire training set (1 epoch).  At test time, the network predicted segmentations at roughly 75 frames per second (real-time).

\subsection{Tracking}

Short axis (\SA{}) scans from \NumO{} subjects with overt hypertrophic cardiomyopathy (\HCM{}) and \NumC{} control subjects were segmented as described above.  For each scan, the \LV{} endocardium, \LV{} epicardium, and \RV{} endocardium were tracked using the spline optimization method.  The tracking algorithm was implemented in the \texttt{C++} programming language using the Insight Toolkit (ITK) for reading, writing, and manipulating images and point sets, and using the Ceres Solver for least squares optimization.  The three registration passes took $\sim 2.1s$ per cine sequence.

\subsection{Regional Function}

Following tracking, registered splines from the \NumSubjectsTotal{} subjects were used to calculate global strain.  For each structure in each \SA{} plane, global strain was compared between \HCM{} and control subjects using the Student's t test.

\section{Results}

In terms of segmentation, performance relative to expert manual segmentation was similar across all networks: mean pixel accuracy was $\sim 97\%$, intersection-over-union (\IoU{}) across all classes was $\sim 87\%$, and \IoU{} across foreground classes only was $\sim 84\%$.  However, inspection of the images revealed that a single subject with severe, nonuniform illumination was incorrectly segmented by all networks, with a disproportionate effect on mean performance metrics.  For this reason, median values are also reported.  Broadly, performance improved with the addition of temporal context over the target frame alone, and with the dilated convolution (\DC{}) networks compared with the \UNet{} networks (Table~\ref{tab:architectureaccuracy}).  Therefore, Network D was selected to provide segmentations for the tracking data.

\begin{table}
\centering
\caption{Network Performance Compared with Expert Manual Segmentations.} \label{tab:architectureaccuracy}
\begin{tabular}{|c|c|c|c|c|c|} \hline
\hline
Name      & Description & Metric & Pixel Accuracy & \IoU{} (All) & \IoU{} (Foreground)    \\
\hline
\hline
Network A & \UNet{}, No Context & Mean & \ChOneAtFAcc{} & \ChOneAtFIoUAll{} & \ChOneAtFIoUFor{}\\
          &                     & Median & \ChOneAtFAccMed{} & \ChOneAtFIoUAllMed{} & \ChOneAtFIoUForMed{} \\
\hline
Network B & \DC{}, No Context   & Mean & \ChOneAtTAcc{} & \ChOneAtTIoUAll{} & \ChOneAtTIoUFor{}\\
          &                     & Median & \ChOneAtTAccMed{} & \ChOneAtTIoUAllMed{} & \ChOneAtTIoUForMed{} \\
\hline
Network C & \UNet{}, Context    & Mean & \ChTwoAtFAcc{} & \ChTwoAtFIoUAll{} & \ChTwoAtFIoUFor{} \\
          &                     & Median & \ChTwoAtFAccMed{} & \ChTwoAtFIoUAllMed{} & \ChTwoAtFIoUForMed{} \\
\hline
Network D & \DC{}, Context      & Mean & \ChTwoAtTAcc{} & \ChTwoAtTIoUAll{} & \ChTwoAtTIoUFor{} \\
          &                     & Median & \ChTwoAtTAccMed{} & \ChTwoAtTIoUAllMed{} & \ChTwoAtTIoUForMed{} \\
\hline
\end{tabular}
\end{table}

\begin{figure}[htb]
\centering
\includegraphics{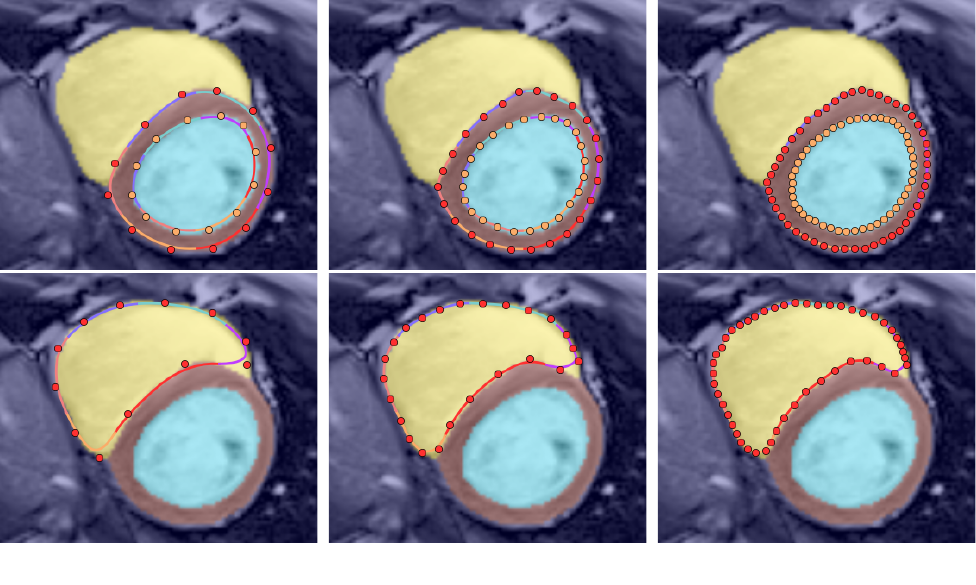}
\caption{Spline registration was conducted in successive passes (left to right), where the output of one was subdivided and used to initialize the next.  \LV{} and \RV{} tracking results are shown above and below, respectively.  Note especially regions of acute curvature, such as the insertion of the \RV{} on the \LV{}, which improves from left to right as the granularity of the spline increases.}
\label{fig:registrationpasses}
\end{figure}

Tracking was performed in three passes (Fig.~\ref{fig:registrationpasses}), where the output of one pass was subdivided and passed to the next pass.  In each pass, the contours tighten towards the segmentation, allowing for acute structures such as the \RV{} insertion points to be better described.  Compared with registering a highly subdivided spline directly, this technique avoids local minima and converges faster.  

The relationship between strain values measured in the control and overt groups was consistent with other studies \cite{Vigneault2014a}.  In particular, circumferential strain measured in the equatorial \LV{} endocardium was higher (more negative) in overt subjects relative to control subjects (\ELVEndO{} vs \ELVEndC{}, $p = $ \ELVEndp{}).  Detailed circumferential strain results are given in Table~\ref{tab:StrainResults}.

\begin{table}
\centering
\caption[Circumferential Strain Results]{Circumferential Strain Results.  ($^\dagger$: Significant at the $p < 0.05$ level.)} \label{tab:StrainResults}
\begin{tabular}{llccc} \hline
\hline
Plane & Structure & Control ($\%$) & Overt ($\%$) & p \\
\hline
\hline
Base     & LV Endocardium & $\BLVEndC{}$ & $\BLVEndO{}$ & $\BLVEndp{}$ \\
         & LV Epicardium  & $\BLVEpiC{}$ & $\BLVEpiO{}$ & $\BLVEpip{}$ \\
         & RV Endocardium & $\BRVEndC{}$ & $\BRVEndO{}$ & $\BRVEndp{}$ \\
Midslice & LV Endocardium & $\ELVEndC{}$ & $\ELVEndO{}$ & $\ELVEndp{}^\dagger$ \\
         & LV Epicardium  & $\ELVEpiC{}$ & $\ELVEpiO{}$ & $\ELVEpip{}$ \\
         & RV Endocardium & $\ERVEndC{}$ & $\ERVEndO{}$ & $\ERVEndp{}$ \\
Apex     & LV Endocardium & $\ALVEndC{}$ & $\ALVEndO{}$ & $\ALVEndp{}$ \\
         & LV Epicardium  & $\ALVEpiC{}$ & $\ALVEpiO{}$ & $\ALVEpip{}$ \\
         & RV Endocardium & $\ARVEndC{}$ & $\ARVEndO{}$ & $\ARVEndp{}$ \\
\hline
\end{tabular}
\end{table}

\begin{figure}[htb]
\begin{center}
\includegraphics{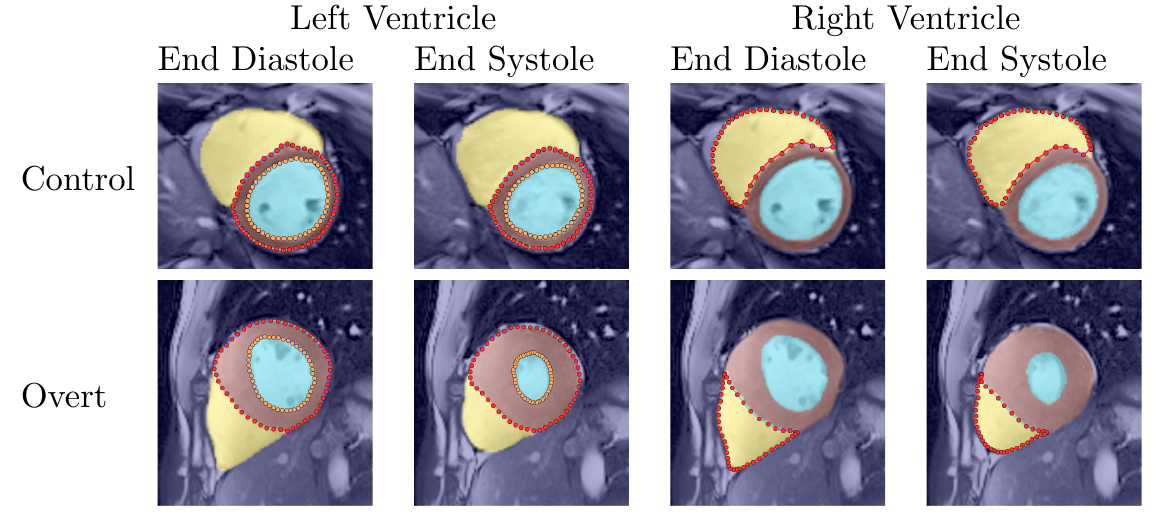}
\end{center}

\caption{Representative segmentation and tracking results in a control subject (top) and a patient with overt \HCM{}.}
\label{fig:SegmentationControlOvert}
\end{figure}

Representative segmentation and tracking results are shown for control and overt subjects (Fig.~\ref{fig:SegmentationControlOvert}).  The model learned to avoid the papillary muscles of the \LV{} myocardium and performed well even in subjects with severe hypertrophy.  Tracking visually followed the contours of the segmentation closely.

\section{Conclusions}

Measuring cardiac function in a fully automated way from \SSFP{} \CMR{} has the potential to simplify quantification of regional cardiac function, and expedite clinical adoption.  We have presented a fully automated pipeline for cardiac segmentation, tracking, and estimation of cardiac strain.  We obtained segmentation results with and without temporal context in \UNet{} and \DC{} architectures, and found improvements with temporal context, as well as in \DC{} architectures.  The best-performing architecture (Network D) had a median pixel accuracy of $\ChTwoAtFAccMed{}$, all-class \IoU{} of $\ChTwoAtFIoUAllMed{}$, and foreground \IoU{} of $\ChTwoAtFIoUForMed{}$.  We then presented a feature tracking algorithm taking these segmentations as input and jointly optimizing a set of quadratic splines over all frames simultaneously.  We applied this segmentation and tracking to the \LV{} endocardium, \LV{} epicardium, and \RV{} endocardium of healthy and disease subjects, and found statistically significant differences between control and overt \HCM{} subjects consistent with previous studies.

Our algorithm is novel in three principal ways.  First, in terms of application, the wide anatomical variability observed in subjects with overt \HCM{} make segmentation a particularly difficult problem; this work is the first to demonstrate that \CNN{}-based segmentation is effective in these subjects.  In addition, we have directly compared dilated convolution and \UNet{} architectures to select an appropriate state-of-the-art architecture solution to this problem.  Second, a persistent problem in \FTCMR{} is the interference of the papillary muscles in cardiac segmentation; we have demonstrated that deep learning neatly solves this problem, and are unaware of deep learning segmentation being used for \FTCMR{} before.  Third, the spline optimization method presented avoids the errors inherent to the various pairwise sequential and reference frame formulations ubiquitous in the feature tracking literature to date.

Notably, all networks failed to segment a single case with severe nonuniform illumination.  Augmentation during training to counteract this effect will be the subject of future work.  Moreover, because only edge features are tracked, our method suffers from the so-called ``aperture-problem,'' such that anatomical correspondence may not be reliable.  In future work, we will incorporate features from our pre-trained \CNN{} into the spline optimization to mitigate this effect.  However, this may be a fundamental limitation of \FTCMR{} where trabeculation is minimal, such as when measuring \LV{} endocardial strain in the basal slice.

In conclusion, we have presented a fully automated pipeline which addresses a number of longstanding challenges to the adoption of \FTCMR{}, and tested this pipeline successfully in the context of a difficult clinical problem.

\section{Acknowledgements}

D. Vigneault is supported by the NIH-Oxford Scholars Program and the NIH Intramural Research Program.  W. Xie is supported by the Google DeepMind Scholarship, and the EPSRC Programme Grant Seebibyte EP/M013774/1.

\bibliographystyle{ieeetr}
\bibliography{citations}

\end{document}